\documentclass[10pt,a5paper,twoside]{article}
\usepackage{nodalida13}
\usepackage[english]{babel}
\usepackage[T1]{fontenc}
\usepackage[utf8]{inputenc}
\usepackage{relsize}
\usepackage{latexsym}
\usepackage{amsmath}
\usepackage{caption}
\usepackage{subcaption}
\title{Properties of phoneme $N$-grams across the world's language families}

\begin{document}
\frenchspacing
\maketitle

\abstract{
In this article, we investigate the properties of phoneme $N$-grams across half of the world's languages. We
investigate if the sizes of three  different $N$-gram distributions of the world's language families obey a power law.
Further, the $N$-gram distributions of language families parallel the sizes of the families, which seem to obey a
power law distribution. The correlation between $N$-gram distributions and language family sizes improves with
increasing values of $N$. We applied statistical tests, originally given by physicists, to test the hypothesis of power
law fit to twelve different datasets. The study also raises some new questions about the use of $N$-gram distributions
in linguistic research, which we answer by running a statistical test.}


\newpage

\section{Introduction and related work}
\subsection{Power laws}
Many real-life phenomena such as word-type frequencies in a corpus, degrees of nodes in a network representation of the
internet, the number of species in a genus of mammals and populations of cities follow a power law
distribution. Power law distributions seem ubiquitous in nature and many other phenomena are also claimed to obey a
power law~\cite{clauset2009power}. Computational linguists will typically have come across power laws in
a form popularly known as Zipf's law \cite{zipf1935}. Zipf's law is stated as $f\propto r^{-1}$, where $f$ and $r$ is the frequency
and rank of a word type $x$. This is a special case of the power law with the probability density function $p(x)$
defined as $p(x)= Cx^{-\alpha}$ where $\alpha = 1$. $\alpha$ is the scaling parameter ($\alpha > 1$) and $C$ is the
normalizing constant. If $p(x)$ is lower-bounded at $x_{min}$ then the power law assumes the form of $p(x) = (\alpha
-1)x_{min}^{\alpha-1}\cdot x^{-\alpha}$.\footnote{$C$ is calculated by solving for $\int_{x_{min}}^{\infty}Cx^{-\alpha} =
1$ for $\forall x \in \mathbb{R}$. In this paper, $x$ takes up integer values only.} Power law is just one member of a
larger class of distributions called large number of rare events (LNRE) distributions~\cite{baayen1991stochastic}. As
pointed by~\citet{evert2007zipfr}, LNRE distributions have applications in NLP/CL. LNRE distributions can be used to
predict the total vocabulary size from a smaller sample. We now turn to a discussion of some recent work in
computational historical linguistics where power laws play a central role in the argumentation.

\begin{figure}[ht!]
\begin{subfigure}[b]{0.45\textwidth}
\centering
\includegraphics[width=\textwidth]{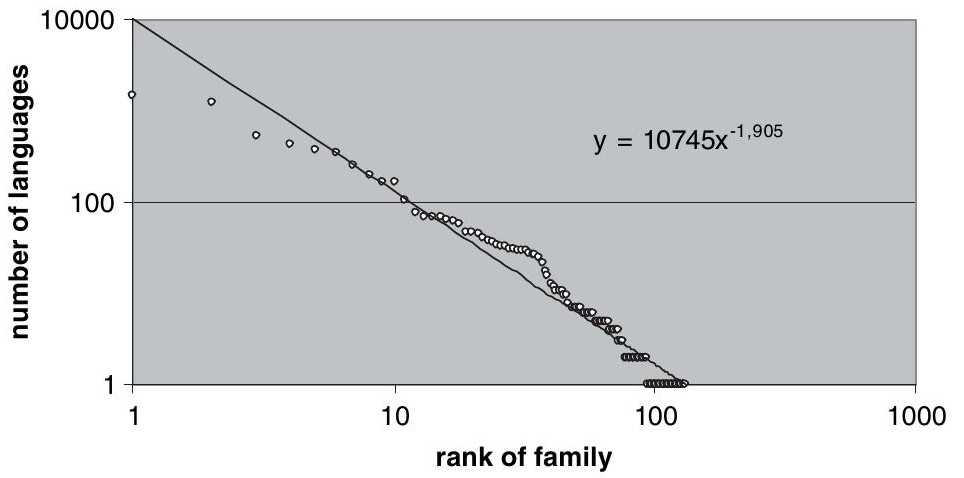}
\caption{\citet{wichmann2005power}.}
\label{fig:figure1}
\end{subfigure}
~
\begin{subfigure}[b]{0.45\textwidth}
\centering
\includegraphics[width=\textwidth]{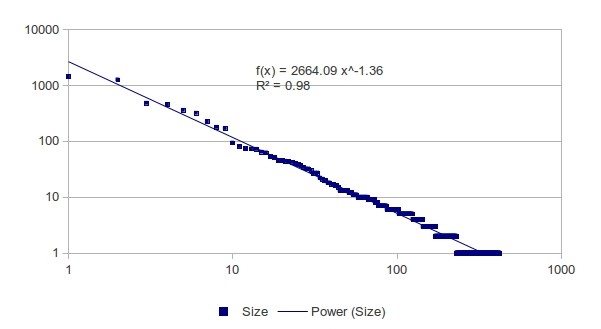}
\caption{\citet{haraldagri}.}
\label{fig:figure2}
\end{subfigure}
\label{fig:powerlaws}
\caption{Frequency-rank plots for two different classifications along with the $r^{2}$ and the regression-lines
generated using a commonly available spreadsheet package.}
\end{figure}

There are about $7000$ languages in the world \cite{ethnologue,haraldagri}, forming more than $120$
families
according to the \emph{Ethnologue}, whereas more than $400$ are listed by \citet{haraldagri}. A language
family is a group of related languages (or a single language when there are no known related languages, such as Basque)
descended from a common ancestor~\cite{campbell2008language}. Each of these language
families is assigned a tree structure in at least two classifications~\cite{ethnologue,haraldagri}. The
size\footnote{\citet{haraldagri} uses \textbf{cardinal size} to indicate family size.} of a language family is defined
as the number of related languages included in the family. \citet{wichmann2005power} observes that the
frequency-rank\footnote{In this context, frequency denotes the family size.} plot of the sizes of language families (as
defined in \emph{Ethnologue}) seems to follow a power law. Figure~\ref{fig:figure1} (reproduced
from~\citealt{wichmann2005power}) is plotted on a log-log scale and shows
a slight deviation from the regression line in the region of $50$ and $\ge 100$. Figure~\ref{fig:figure2} shows the
frequency-rank plot for~\citeauthor{haraldagri}'s classification.

There is a slight deviation from the strict adherence to the straight line in
Figure~\ref{fig:figure2}. However, the goodness-of-fit $r^{2}$ is in the range of $0.957$ and $0.98$ in both the
classifications. Looking into the closely related field of linguistic typology,~\citet{maslova2008meta} proposes that
\emph{meta-typological} distributions obey power law. A meta-typological distribution is defined as the
number of languages having a particular linguistic feature value, such as a particular word order or a phoneme inventory 
of a particular size (e.g., small, medium, or large).\footnote{The data for these
experiments is derived from the \emph{World Atlas of Language Structures} (WALS;~\citealt{haspelmath2008wals}). The data is
generated by random selection of a linguistic feature value and counting the number of languages for that value.} In
response,~\citet{Cysouw:2007:PDT:1886644.1886647} proposes that the distribution is actually exponential masquerading as a power
law.

\subsection{Testing power laws}
The scaling parameter $\alpha_{sp}$, $\alpha$ -- estimated using a spreadsheet package -- in Figures~\ref{fig:figure1}
and~\ref{fig:figure2} is $1.905$ and $1.36$ respectively. Apart from the high $r^{2}$ value, there seems to be no
independent statistical test for the support of a power law. However, a recent paper by~\citet{clauset2009power}
revisited this topic and proposed a number of statistical tests for validating power law models. The authors provide a
maximum likelihood estimate of the two parameters, $x_{min}$ and $\alpha$ and a method for computing the statistical
significance score of the estimates. Further, they test the superiority of the power law with respect to candidate
distributions, listed in Table~\ref{tab:table1}.

In a recent paper,~\citet{jager2012power} applied the statistical tests of~\citet{clauset2009power} to test the
fit of the power law model to global linguistic datasets such as frequency of color terms, phonological templates
for selected basic vocabulary items, and meta-typological distributions.~\citeauthor{jager2012power} also applied a
series of statistical tests to \citeauthor{maslova2008meta}'s data and showed that a power law with exponential cutoff
describes the data better than a power-law model.

\begin{table}[ht!]
\center
 \begin{tabular}{|p{3cm}|c|p{2cm}|}
\hline
  name & probability density function $p(x)$ & \# of parameters ($m$)\\
\hline
  power law (PL)& $(\alpha-1)x_{min}^{\alpha-1}x^{-\alpha}$ & $1$\\
  power law with exponential cutoff (PLWC)& $\frac{\lambda^{1-\alpha}}{\Gamma (1-\alpha,\lambda x_{min})}
x^{-\alpha}e^{-\lambda x} $ & $2$\\
  log-normal (LN)& $ \sqrt{\frac{2}{\pi\sigma^2}}\left [ \operatorname{erfc}\left ( \frac{\ln x_{min} -
\mu}{\sqrt{2}\sigma} \right ) \right ]^{-1} \frac{1}{x}\exp\left[{-\frac{(\ln x-\mu)^2}{2\sigma^2}}\right]$ & $2$\\
  exponential ($\exp$)& $\lambda e^{\lambda x_{min}} e^{-\lambda x}$ & $1$\\
  stretched exponential (str $\exp$)&$\beta\lambda e^{\lambda x_{min}^{\beta}} x^{\beta - 1} e^{-\lambda x^{\beta}}$ &
$2$\\
  gamma ($\Gamma$)& $\frac{1}{\Gamma(k)\theta^k}x^{k-1}e^{-x/\theta}$ & $2$\\
\hline
 \end{tabular}
\caption{List of various candidate distributions and the number of parameters in each model. These distributions are
popularly referred to as ``heavy-tail'' distributions.}
\label{tab:table1}
\end{table}

The standard method for testing a power law hypothesis consists of plotting a frequency-rank plot on a log-log scale
and applying a linear regression. The linear regression boils down to determining the parameters of $\log p(x) = c +
\alpha \log(x)$. However,~\citet{clauset2009power} warn against this. Further, they demonstrate that the value of
estimated $\alpha$ differs largely from that derived from the regression analysis. The validity of the power law is
tested through the following steps:
\begin{itemize}
 \item Estimate $\alpha_{sp}$ and $r^2$ using a spreadsheet package by plotting the frequency-rank plot of the data on
a log-log scale.
 \item Estimate $\alpha_{est}$ and $x_{min}$ based on the maximum likelihood criterion
($L$).
\item The preference of a power law to rest of the candidate distributions is tested through a likelihood ratio
test~\cite{dunning94cl} under a significance criterion of $p\le0.1$.
\item The absolute goodness-of-fit of a model is computed using the Akaike Information
criterion which is defined as in~\ref{eq:aic}, $m$ is the number of parameters and $L$ is the
goodness-of-fit. The model with lowest AIC is the best fit.

\begin{equation} \label{eq:aic}
 AIC = 2m - 2\log(L)
\end{equation}
\end{itemize}

\citeauthor{jager2012power} simplifies the computation of $\alpha$ for discrete data by assuming a continuous
approximation of the power-law model and fixing $x_{min}$ at $0.5$. This assumption implies that all data points in a
dataset completely fit a power-law model. However, it can always be the case that only a part of the dataset follows a
power-law model. As a necessary diversion, it is useful to know the computation of $x_{min}$. The scaling parameter
$\alpha$ is estimated by successive removal of the lowest value of $x$. The fitted distribution is then compared to the
empirical distribution through a Kolmogorov-Smirnov statistic ($D$). The value of $x$ which minimizes $D$ is chosen as
$x_{min}$.

In this paper, we find that the rank plot of the size of phoneme $N$-grams for 45 language families seems to obey a
power law distribution as given in Figure~\ref{fig:figure3}. This finding is in parallel to that
of~\citet{wichmann2005power}. By applying the statistical procedure mentioned above, we attempt to establish whether the
family
sizes given in three different language classifications actually obey a power law. Subsequently we test if the phoneme $N$-grams
also obey a power law model. We describe the database in the next section.

\section{Database}\label{sec:database}
\begin{figure}[ht!]
\begin{subfigure}[b]{0.45\textwidth}
\centering
\includegraphics[width=\textwidth]{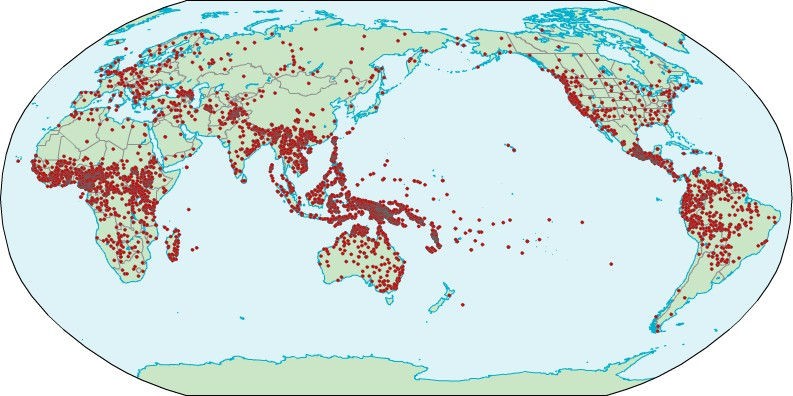}
\caption{World map}
\label{fig:figure3}
\end{subfigure}
~
\begin{subfigure}[b]{0.45\textwidth}
\centering
\includegraphics[width=\textwidth]{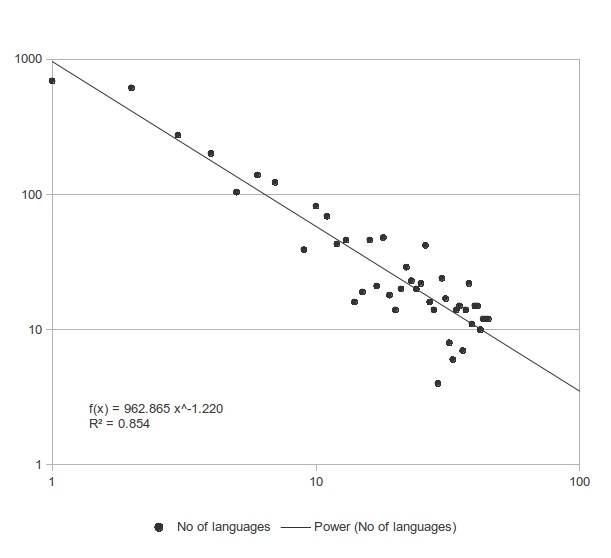}
\caption{Frequency-rank plots}
\label{fig:figure4}
\end{subfigure}
\caption{ASJP word lists on world map and the plots with $r^2$ values for families from Ethnologue.}
\label{fig:asjp}
\end{figure}

In this section, we describe the global linguistic database (depicted in Figure~\ref{fig:figure3}). A consortium of
international scholars known as ASJP (Automated Similarity Judgment Program;~\citealt{wichmannasjp14})
have collected reduced word lists -- $40$ items from the original $200$ item Swadesh word lists \cite{swadesh1955towards}, selected for maximal
diachronic stability -- for more than half of the world's languages and embarked on an ambitious program for
investigating automated classification of the world's languages. The ASJP database in many cases includes more than one
word list for different varieties of a language (identified through its ISO 693-3 code). A word list is included into the
database if it has attestations of at least $28$ of the $40$ items ($70\%$).

Only language families with at least $4$ members are included in our experiments. This leaves a dataset with $45$
language families representing $3151$ languages (or $4524$ word lists) of the world. The names of the language families
are as defined in the \emph{Ethnologue}. A word list might include known borrowings marked as such and
these are not used in our experiments. The words in the ASJP database are transcribed using a reduced
phonetic transcription known as \emph{ASJP code} consisting of $34$ consonants, $7$ vowels and a symbol for
nasalization, and two other `modifiers', which are used to indicate that preceding symbols combine as single segments.
All click sounds are reduced to a single click symbol and distinctions such as tones, vowel length, and stress are
ignored. The computation of a phoneme $N$-gram profile for a language family is described in section~\ref{sec:exps}.
The frequency-rank plot of the language families in the current sample is shown in Figure~\ref{fig:figure4}. The
regression shows a $r^2$ value of $0.85$ which is quite high.

\section{Experiments and results}\label{sec:exps}
\begin{figure}[h]
\centering
\includegraphics[height=0.45\textwidth]{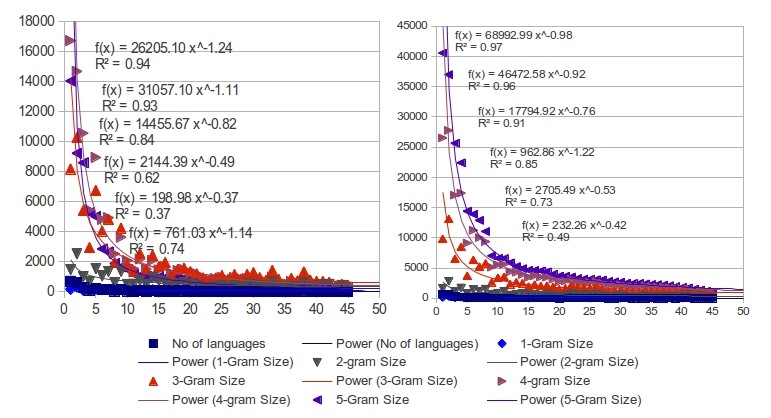}
\caption{N-gram profiles fitted to a power law.}
\label{fig:figure5}
\end{figure}

All the word lists belonging to a single language family are merged together. Recall that the ASJP database can include
more than one word list -- representing different varieties -- for a language. All the consecutive symbol sequences of
length varying from $1$--$5$ are extracted and the size of the $N$-gram profile is defined as the total number of unique
$1$--$N$-grams obtained through this procedure. Thus, a $3$-gram profile consists of all the phoneme $1$-, $2$- and
$3$-grams. The size of the $3$-, $4$- and $5$-gram profiles for each of the language families as defined in the
\emph{Ethnologue} is given in Table~\ref{tab:sizes}. In effect, an $N$-gram profile is the sum of all the $n$-gram types
leading up to $N$. As evident from right panel of Figure~\ref{fig:figure5}, each of the $N$-gram profiles, $N\ge3$, seem
to follow a power law.

When a power law regression is applied to each of the frequency-rank plots, the goodness-of-fit $r^2$ is
$0.91$, $0.96$ and $0.97$ for $3$-grams, $4$-grams and $5$-grams respectively. The $r^2$ value of both $1$-grams and
$2$-grams is quite low, only $0.49$ and $0.73$ when compared to the $r^2$ value of the number of languages, $0.85$. We
also plot the frequency-rank plots for each $n$-gram type in the left panel of Figure~\ref{fig:figure5}. The $r^2$
values are quite high and are $0.84$, $0.93$ and $0.94$ respectively. The $r^2$ scores in Figure~\ref{fig:figure1} for
$4$--$5$ grams are very high and fall within the range of the correlation of $0.957$ (with language family size),
reported by~\citealt{wichmann2005power}.

\begin{table}[h!]
\center
\scriptsize
\begin{tabular}{|lrrrr|lrrrr|}
\hline
\textbf{Language family} & \textbf{NOL} & \textbf{3-gram} & \textbf{4-gram} &
\textbf{5-gram} & \textbf{Language family} & \textbf{NOL} & \textbf{3-gram} &
\textbf{4-gram} & \textbf{5-gram} \\ \hline
Austronesian & 692 & 9808 & 26527 & 40561 & Macro-Ge & 20 & 1813 & 2684 & 3180 \\
Niger-Congo & 615 & 13085 & 27766 & 36987 & Sepik & 22 & 1677 & 2623 & 3144 \\
Trans-NewGuinea & 275 & 6492 & 17051 & 25634 & Tai-Kadai & 42 & 2036 & 2723 & 3077 \\
Afro-Asiatic & 201 & 8456 & 17403 & 22403 & Chibchan & 16 & 1522 & 2484 & 3035 \\
Australian & 104 & 3690 & 9143 & 14401 & WestPapuan & 14 & 1244 & 2203 & 2806 \\
Indo-European & 139 & 6320 & 11252 & 13896 & EasternTrans-Fly & 4 & 1300 & 2166 & 2729 \\
Nilo-Saharan & 123 & 5224 & 10046 & 12891 & Dravidian & 24 & 1376 & 2238 & 2674 \\
Sino-Tibetan & 137 & 5753 & 9386 & 11043 & LakesPlain & 17 & 1142 & 2010 & 2518 \\
Arawakan & 39 & 2626 & 5148 & 7021 & Border & 8 & 1132 & 1922 & 2467 \\
Austro-Asiatic & 82 & 3370 & 5552 & 6608 & South-CentralPapuan & 6 & 1074 & 1878 & 2464 \\
Oto-Manguean & 69 & 3522 & 5607 & 6579 & Penutian & 14 & 1326 & 2017 & 2384 \\
Uto-Aztecan & 43 & 2318 & 4395 & 5873 & Panoan & 15 & 1192 & 1915 & 2288 \\
Altaic & 46 & 2634 & 4304 & 5248 & Witotoan & 7 & 1185 & 1847 & 2264 \\
Salishan & 16 & 2492 & 3944 & 4903 & Hokan & 14 & 1253 & 1864 & 2192 \\
Algic & 19 & 1922 & 3466 & 4643 & Quechuan & 22 & 1101 & 1734 & 2093 \\
Tupi & 46 & 2250 & 3722 & 4619 & Siouan & 11 & 1131 & 1674 & 1952 \\
Torricelli & 21 & 2011 & 3518 & 4523 & Na-Dene & 15 & 1225 & 1637 & 1810 \\
Mayan & 48 & 2083 & 3485 & 4386 & Hmong-Mien & 15 & 1246 & 1563 & 1717 \\
Tucanoan & 18 & 1880 & 3162 & 3979 & Totonacan & 10 & 679 & 1139 & 1510 \\
Ramu-LowerSepik & 14 & 1491 & 2738 & 3676 & Khoisan & 12 & 995 & 1265 & 1377 \\
Carib & 20 & 1662 & 2868 & 3649 & Sko & 12 & 775 & 1068 & 1179 \\
NorthCaucasian & 29 & 2180 & 3158 & 3537 & Mixe-Zoque & 12 & 625 & 897 & 1028 \\
Uralic & 23 & 1896 & 2818 & 3284 & & & & &\\ \hline
\end{tabular}
\caption{The number of languages (NOL), 3-gram, 4-gram and 5-gram profiles for 45 language families.}
\label{tab:sizes}
\end{table}

As we have shown above, the correlation of $N$-gram distribution to language family size improves with increasing $N$
(for $N=1-5$). This is a kind of behavior familiar from corpus studies of word distributions \cite{baayen}, where
closed-class items -- typically function words -- yield distributions similar to the $1$-grams (phonemes) in this study,
whereas open-class words display typical power-law behavior for all corpus sizes, just like the $3$--$5$-grams
in this study. We take this as an indication that we are on the right track, investigating a genuine linguistic
phenomenon. We also test if the genus size across the world's languages displays a power-law like behavior. A \emph{genus} (pl. \emph{genera}) is a
language classification unit which contains related languages and is estimated to be 3000--3500 years
old. The genus level was originally introduced by \citet{dryer2000counting}. We use the genus information given in ASJP database.
The current dataset has 538 genera and 5315 word lists. Table~\ref{tab:aic} shows the results of the application
of the statistical tests to different datasets.

\begin{table}[h!]
\centering
\scriptsize
\begin{tabular}{|p{1.5cm}|p{0.5cm}cp{0.5cm}p{0.4cm}cp{0.4cm}p{0.7cm}p{0.7cm}p{0.7cm}p{0.7cm}p{0.8cm}|}
\hline
Data & $x_{min}$ & $\ln(L)$ & $n_{tail}$ &
$\alpha_{est}$ & PL & $\alpha\_{sp}$ & PLWC & LN & $\exp$
& str $\exp$ & $\Gamma$ \\\hline
\citeauthor{haraldagri} & 1 & -1040.715 & 423 & 1.667 & 1041.715 & 1.36 & 1040.144 & 1059.037 & 1613.822 &
\textbf{1039.84} &1090.109 \\
ASJP & 7 & -211.167 & 43 & 1.633 & 212.167 & 1.22 & 209.791 & \textbf{209.137} & 224.509 & 209.409 & 210.587 \\
WALS genus & 5 & -720.257 & 193 & 1.898 & 721.257 & 1.26 & 714.354 & 714.95 & 747.291 & 714.197 & \textbf{708.282} \\
1-grams & 51 & -166.455 & 35 & 2.965 & 167.455 & 0.37 & 167.193 & 168.004 & 167.56 & \textbf{167.193} & 167.43 \\
2-grams & 367 & -239.206 & 35 & 2.844 & 240.206 & 0.49 & \textbf{239.892} & 241.277 & 241.34 & 239.92 & 241.153 \\
3-grams & 762 & -303.22 & 37 & 2.262 & 304.22 & 0.82 & \textbf{303.293} & 305.221 & 308.871 & 303.323 & 306.022 \\
4-grams & 270 & -387.603 & 45 & 1.638 & 388.603 & 1.11 & 383.994 & \textbf{382.972}& 390.789 & 383.436 & 384.73\\
5-grams & 173 & -340.531 & 41 & 1.605 & 341.531 & 1.24 & 337.376 & \textbf{335.605} & 343.885 & 336.558 & 337.789 \\
2-grams\textdagger & 506 & -188.941 & 27 & 3.047 & 189.941 & 0.53 & \textbf{189.894} & 191.251 & 190.82 & 189.934 &
191.04 \\
3-grams \textdagger& 1074 & -342.859 & 41 & 2.395 & 343.859 & 0.76 & \textbf{343.198} & 345.588 & 348.941 & 343.245 &
346.15 \\
4-grams\textdagger & 1734 & -346.368 & 38 & 2.196 & 347.368 & 0.92 & \textbf{346.475} & 348.748 & 354.397 & 346.507 &
350.041 \\
5-grams \textdagger& 2093 & -357.069 & 38 & 2.137 & 358.069 & 0.98 & \textbf{357.296} & 359.594 & 366.483 & 357.339 &
361.106 \\
\hline
\end{tabular}
\caption{The first three rows correspond to the language size data of~\citet{haraldagri}, ASJP data (from
Table~\ref{tab:sizes}) and genera sizes. Columns 2--5 correspond to the estimated parameters in a power law. Column 5
shows the AIC value for a power law. Column 6 shows the $\alpha_{sp}$ estimated by a standard spreadsheet package. The
remaining columns correspond to the AIC values for the other candidate distributions. The last four rows show the fit
of each $n$-gram profile leading upto $5$ and are indicated by a \textdagger. For each dataset, the best fit model is
indicated in \textbf{bold}. Here, the common factor $2$ is not included in the AIC computation. All the above results
are computed using the power-law python package~\cite{powerlaws}.}
\label{tab:aic}
\end{table}
Judging by AIC, none of the classification unit datasets follow a power-law distribution. It is important to notice
that the $\alpha_{est}$ widely differs from $\alpha_{sp}$. As demonstrated by~\citet{clauset2009power}, there can be a
large difference when estimating $\alpha$ for small datasets of size $\le50$. Only the $2-5$-gram profiles follow
a power-law with cutoff model ascertained by the lowest AIC value. Interestingly, $n_{tail}$ values are highest for
$n=3$ followed by $4$ and $5$. The value of $\alpha$ for a power law is typically between 2 and 3. The values of
$\alpha_{est}$ for $N$-gram profile also lie between 2 and 3.

\begin{table}[h!]
\small
\centering
\begin{tabular}{|l|ccccc|}
\hline
Data & PLWC & LN & $\exp$ & str $\exp$ & $\Gamma$ \\\hline
\citeauthor{haraldagri} & $-$\textbf{0.0} & \textbf{0.046} & \textbf{0.0} & $-$\textbf{0.005} & \textbf{0.087} \\
ASJP & $-$\textbf{0.012} & $-$\textbf{0.006} & 0.101 & $-$\textbf{0.004} & $-$0.31 \\
genus & $-$\textbf{0.018} & $-$\textbf{0.065} & 0.129 & $-$\textbf{0.009} & $-$\textbf{0.063} \\
1-grams & $-$\textbf{0.097} & $-$0.802 & 0.962 & $-$0.104 & $-$0.48 \\
2-grams & $-$0.109 & 0.969 & 0.613 & $-$0.133 & $-$0.972 \\
3-grams & $-$\textbf{0.086} & 0.999 & 0.157 & $-$\textbf{0.099} & 0.684 \\
4-grams & $-$\textbf{0.002} & $-$\textbf{0.003} & 0.7 & $-$\textbf{0.001} & $-$\textbf{0.021} \\
5-grams & $-$\textbf{0.001} & $-$\textbf{0.001} & 0.668 & $-$\textbf{0.0} & $-$\textbf{0.011} \\
2-grams\textdagger & $-$0.168 & 0.849 & 0.619 & $-$0.194 & 0.936 \\
3-grams\textdagger & $-$\textbf{0.093} & 0.727 & 0.13 & $-$\textbf{0.097} & 0.53 \\
4-grams\textdagger & $-$\textbf{0.096} & 0.851 & \textbf{0.056} & $-$\textbf{0.097} & 0.43 \\
5-grams\textdagger & $-$\textbf{0.099} & 0.789 & \textbf{0.035} & $-$\textbf{0.09} & 0.357 \\\hline
\end{tabular}
\caption{The table shows the results of the likelihood-ratio test for comparing the power law with the rest of candidate
distributions. The $-$ sign indicates the test favoring the candidate model than the power law model. Each number is
the $p$-value and the significance is indicated in \textbf{bold}.}
\label{tab:llr}
\end{table}

The AIC values in Table~\ref{tab:aic} suggest that the power-law with cutoff is a better model than power-law for for
$N$-gram profiles. We assess this superiority through a likelihood ratio test. The results are given in
Table~\ref{tab:llr}. The results suggest that the PLWC is a better fit than PL at a significance criterion $p\le0.1$.
Interestingly, none of the family-size datasets are genuinely power-lawish. They seem to belong to other
``heavy-tailed'' distributions. Incidentally,~\citeauthor{haraldagri}'s dataset -- covering more than 7000 languages --
fits better to a stretched exponential model than a power-law distribution.

\begin{figure}[h!]
\begin{subfigure}[b]{0.45\textwidth}
\centering
\includegraphics[height=0.65\textwidth,width=\textwidth]{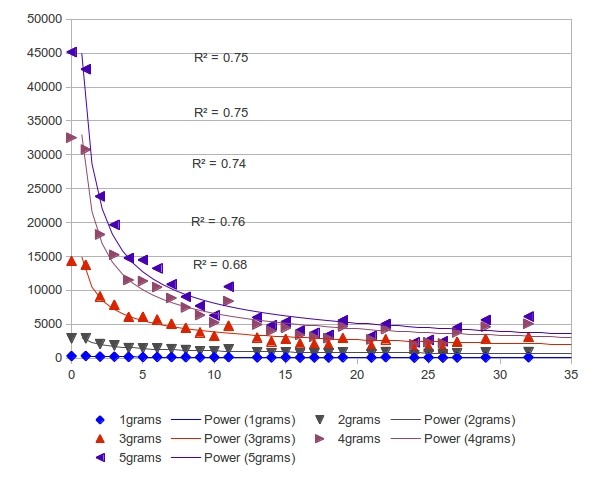}
\caption{Random samples.}
\label{fig:figure6}
\end{subfigure}~
\begin{subfigure}[b]{0.45\textwidth}
\centering
\includegraphics[width=\textwidth,height=0.65\textwidth]{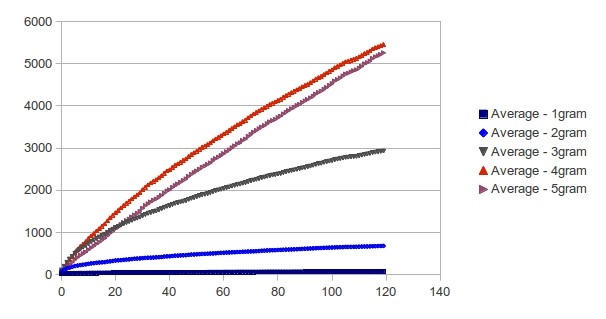}
\caption{Average $N$-gram profile size}
\label{fig:figure7}
\end{subfigure} 
\end{figure}

Even though this study shows that phoneme $N$-gram profiles closely mirror the power-law-with-cutoff behavior, it
raises more questions than it answers about the use of $N$-gram distributions in linguistic research, such as:
\begin{itemize}
\item[Q.] Is the $N$-gram distribution an effect strictly connected with genetic relatedness among the languages, or
simply an effect of the number of languages in a group (regardless of whether they are related or not)?
\item[A.] We answer this question through the following procedure:
 \begin{enumerate}
\item For a family size $s$, make a random sample of languages of size $s$.
  \item Compute the $N$-gram profiles.
 \end{enumerate}
 
Repeat steps $1-2$ for all family sizes and plot the $N$-gram profile sizes. The results are shown in
Figure~\ref{fig:figure6}. All the $r^2$ values are in the range of $0.68$ to $0.75$. This experiment suggests that the
$N$-gram distribution is related to genetic relatedness and not an effect of a sample size.

\item[Q.] If the effect is genetic, can the size of the family be predicted from $N$-gram profiles of smaller samples
than the full family? (This could be very useful.)
\item[A.] We answer this question through the following procedure:
\begin{enumerate}
 \item For a family of size $s$, create a random language sample of size $i$, where $1\le i \le s$.
 \item Compute the $N$-gram profiles for each random sample.
\item Repeat steps $1-2$ for $10$ iterations and compute the average size of a $N$-gram profile for each $i$.
\end{enumerate}
Repeat the steps $1-3$ for all $i$. The results of this experiment for $s=104$ (Australian family) is shown in
Figure~\ref{fig:figure7}. Figure~\ref{fig:figure7} shows the plot for the average number of $N$-gram types vs. the size
of language sample. All $N$-gram curves (except $1$ and $2$) seems to be increasing monotonically and not
stabilizing after a particular sample size. Only $1$-grams and $2$-grams tend to stabilize with respect to sample
size. The $N$-gram curves for other language families also follow the same trend. These results suggest that the
$N$-grams ($N\ge3$) of smaller samples cannot be used to predict the full family size.
\end{itemize}

\section{Conclusion}
In this paper, we tested if the language units of the three classifications obey a power law. We find that 
the three datasets are not well modeled by a power law model. We then tested if the $N$-gram profiles follow a power law
and observed that they actually follow a power-law with cutoff distribution. Finally, we posed two questions about the
utility of $N$-grams for historical linguistics and found that $N$-grams do not pass the test.

\newpage
\bibliographystyle{apalike}
\bibliography{myreflnks2}

\end{document}